# Improved YOLOv5s model for key components detection of power transmission lines

**Chen Chen[1], Guowu Yuan[1,*], Hao Zhou[1], and Yi Ma[2]**

[1] School of Information Science and Engineering, Yunnan University, Kunming 650504, Yunnan, China
[2] Electric Power Research Institute, Yunnan Power Grid Co., Ltd, Kunming 650214, Yunnan, China

* **Correspondence:** Email: gwyuan@ynu.edu.cn; Tel: +8687165033748.

**Abstract:** High-voltage transmission lines are located far from the road, resulting in inconvenient inspection work and rising maintenance costs. Intelligent inspection of power transmission lines has become increasingly important. However, subsequent intelligent inspection relies on accurately detecting various key components. Due to the low detection accuracy of key components in transmission line image inspection, this paper proposed an improved object detection model based on the YOLOv5s (You Only Look Once Version 5 Small) model to improve the detection accuracy of key components of transmission lines. According to the characteristics of the power grid inspection image, we first modify the distance measurement in the k-means clustering to improve the anchor matching of the YOLOv5s model. Then, we add the convolutional block attention module (CBAM) attention mechanism to the backbone network to improve accuracy. Finally, we apply the focal loss function to reduce the impact of class imbalance. Our improved method's mAP (mean average precision) reached 98.1%, the precision reached 97.5%, the recall reached 94.4% and the detection rate reached 84.8 FPS (frames per second). The experimental results show that our improved model improves the detection accuracy and has advantages over other models in performance.

## 1. Introduction

Inspecting power transmission lines is a daily task in power grid companies. There are increasingly long-distance transmission lines in China. Patrol inspection is very heavy, and operation and maintenance costs are high. Long-distance transmission lines are located in complex terrain and cover a wide range. Manual inspection of key components has been unable to meet the actual needs, and more intelligent detection methods are urgently needed. A large number of transmission line images were taken by cameras installed on transmission lines or unmanned aerial vehicles. These images are used for power transmission line inspection based on computer vision. With the advantages of high efficiency, accuracy and security, the inspection based on images has gradually become an important way of transmission line inspection [1]. For the automatic inspection based on images, it is important to detect the key components of transmission lines accurately. The results of this work are good foundations for detecting key component defects and abnormal events.

Object detection, one of computer vision's most fundamental and challenging problems, has received significant attention in recent years [2]. Due to the tremendous successes of image classification based on deep learning, object detection techniques using deep learning have been actively studied in recent years [3]. In recent years, it has been found that adding an attention mechanism to the detection framework can improve target detection accuracy [4]. The current research also extends to detecting objects from video [5]. After detecting a target, many applications still need to track the target. Target tracking should solve the interference of target occlusion and complex background [6]. Object detection based on deep learning has dramatically affected many applications [7,8].

Some researchers have proposed automatic detection methods for key components of transmission lines. Lin et al. [9] improved the faster region-based convolution network (Faster R-CNN) model to achieve multi-object detection in transmission line patrol images. Although the accuracy has been improved, the detecting speed still needs to be improved. Li et al. [10] improved the single shot multibox detector (SSD) model to detect pin defects in transmission lines. The detection accuracy is higher than that of traditional algorithms, but the recall rate and AP value are not high. Zhang et al. [11] replaced the backbone network with re-parameterization visual geometry group (RepVGG) modules in the You Only Look Once Version 3 (YOLOv3) model and added a multi-scale detection box to the network. The improvement achieved foreign object detection on transmission lines. However, the detection speed still needs to be improved, although the accuracy has been improved. Guo et al. [12] used convolutional neural networks to learn the properties of insulators and to locate them. This approach can efficiently and automatically detect insulators in UAV images, but the accuracy is not sufficient. Nguyen et al. [13] used data enhancement and multi-stage component detection methods to solve training data deficiency, sample imbalance and small target detection. However, the improvement suffers from insufficient detection accuracy. Chen et al. [14] combined YOLOv3 with Super Resolution Convolutional Network (SRCNN), and the accuracy can be higher than Faster R-CNN and SSD by 1–3%. The improvement can achieve almost real-time, but there are still problems in small object detection. Liang et al. [15] detected grid component defects based on the Faster R-CNN model. The method effectively improves the detection accuracy, but there are false positives and false negatives in certain defect detection in this method. Ni et al. [16] also improved the Faster R-CNN model, and they used Inception-ResNet-V2 as a basic feature extraction network, which effectively improved the network operation efficiency. Although the accuracy of transmission line fault is 98.65%, the detection speed needs to be improved. Chen et al. [17] combined deformable convolution network (DCN) and feature pyramid network (FPN), and used a data-driven iterative learning algorithm to form an intelligent closed-loop image processing. This method provides new ideas for improving the efficiency of grid detection, but the detection accuracy needs to be enhanced. Liu et al. [18] constructed a large dataset for transmission lines to better suit the complex detection environment. Their method improves the performance of small objects, but the algorithm has high loss fluctuation and slow convergence.

Because the transmission line images have many tiny targets, significant changes in object scale and shooting angle, complex background, and the balance between detection accuracy and speed need to be taken into account at the same time, a robust and fast model is urgently needed [19–21]. The anchor box of the YOLOv5 model is adaptively generated and can predict at multiple scales. The YOLOv5 model is more suitable for our application than other models, and it can balance speed and accuracy better than the other models, such as Faster R-CNN, SSD, CenterNet, YOLOv3, and YOLOv4 (You Only Look Once Version 4). Therefore, we proposed an improved YOLOv5s model for detecting key targets in power transmission lines.

In this paper, we propose a target detection model for key components of transmission lines according to the requirements of intelligent inspection of power grids. Our model has higher detection accuracy than other models, and the detection speed is also faster than most models. Our model can be integrated into the front-end image acquisition equipment for power grid inspection and monitoring.

## 2. Dataset and its statistics

Our dataset is from Yunnan Limited Company of China Southern Power Grid. The images are captured by the surveillance cameras installed on transmission towers or unmanned aerial vehicles (UAV) for transmission line inspection. Currently, the power grid monitoring system is relatively complete, and the primary image data comes from the surveillance camera. UAVs are only used in a few places where the surveillance camera cannot cover. Therefore, most of the images in the dataset are captured by surveillance cameras, and UAVs capture very few. The dataset contains 4290 images with a resolution of 1920*1080 pixels. The background of these images is complex, including hills, land, and streets.

We have defined and annotated five types of detected targets according to the application needs of power grid companies, and they are screws, poles, insulators, transmission towers, and vehicles. The first four are key components of transmission lines. As huge trucks or engineering vehicles easily damage transmission lines, power grid companies especially require that vehicles are detected objects.

The five categories of labeled targets are shown in Figure 1.

After labeling the dataset, we counted the number and size of various targets, and the results are shown in Figure 2.

Figure 2(a) shows that the width and height of most detected targets only are less than 10% of the image size, and they are small targets. In Figure 2(b), the number of samples in the dataset is not balanced, among which there are fewer vehicles. The dataset shows a large gap between the height and width of the electric pole and the transmission tower.

In response to this dataset's application requirements and characteristics, we will use the YOLOv5s network and improve it for object detection of critical components of power grid transmission lines.

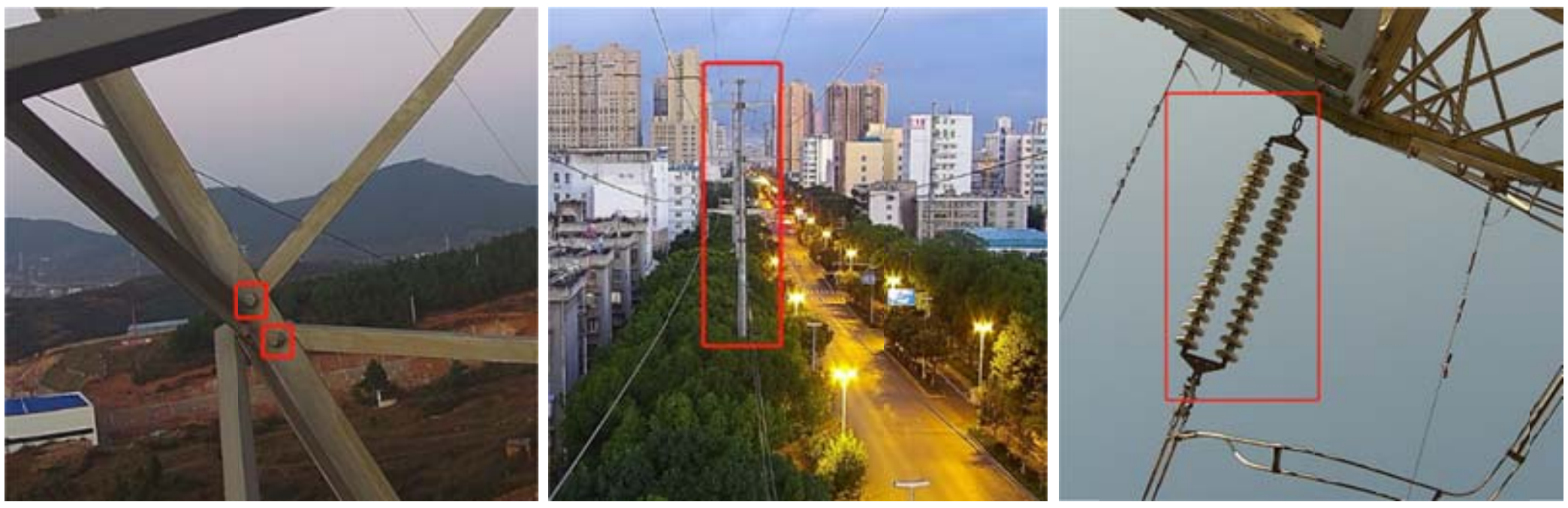

(a) Screws (b) Poles (c) Insulators

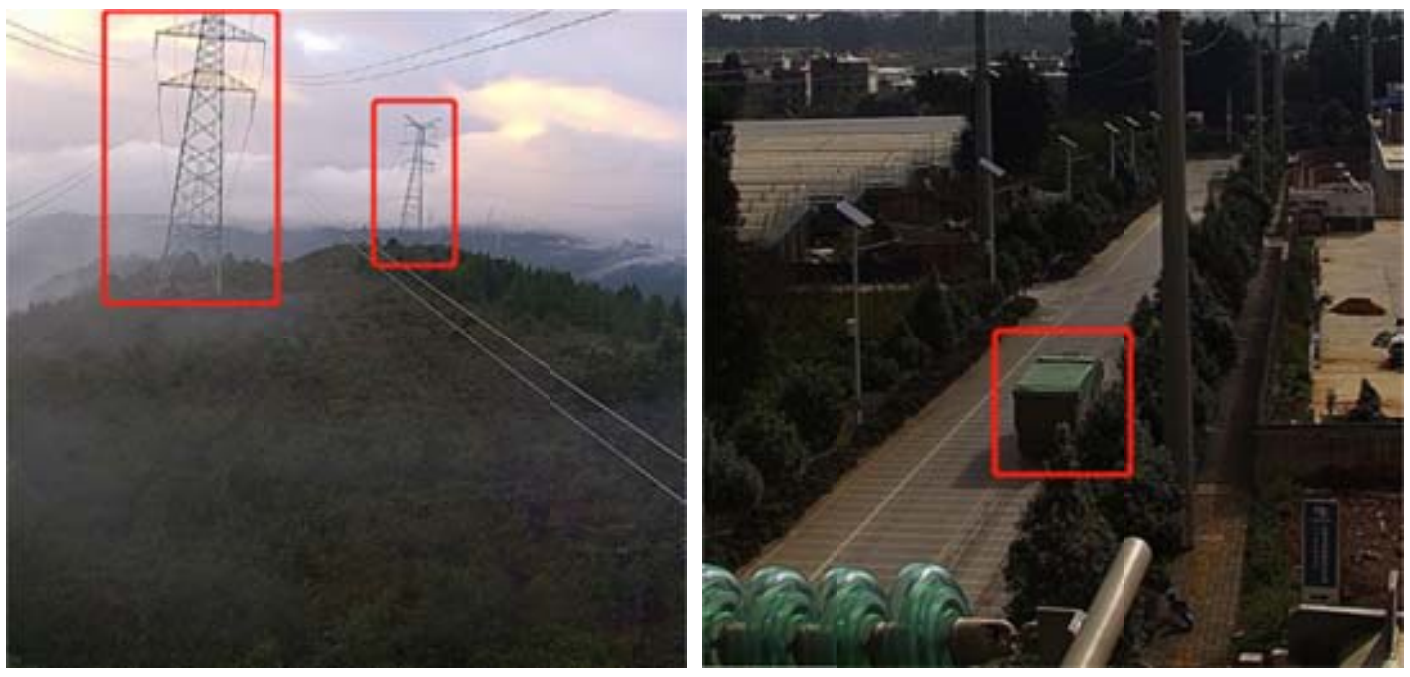

(d) Transmission towers (e) Vehicles

**Figure 1.** Five categories of labeled targets.

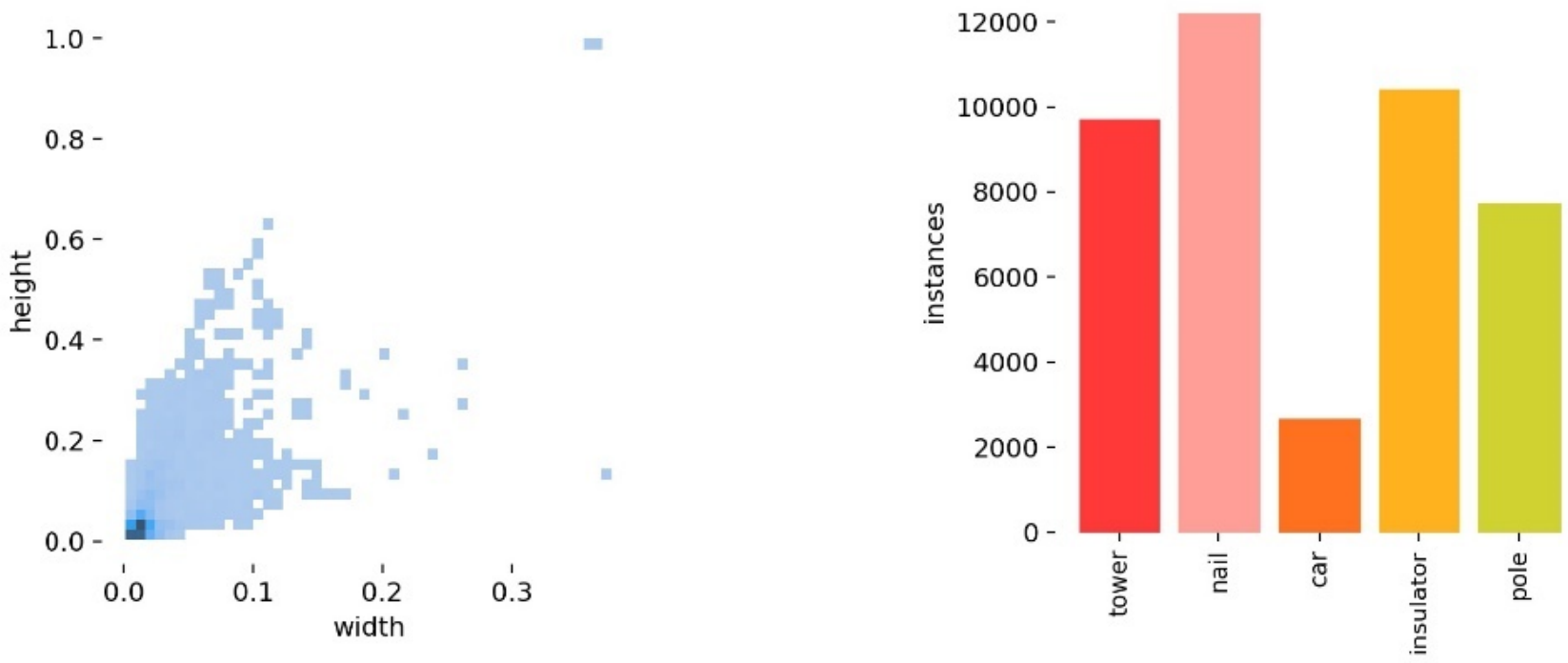

(a) Width and height of detected target (b) Number of various detection targets

**Figure 2.** Statistical results of various targets in the dataset.

## 3. Methods

In practical applications, in the key areas (substations and abnormal monitoring key areas), we consider integrating the detection module into the cameras for real-time detection. In non-key areas (field transmission lines), we capture an image every 30 minutes, transmit it to the data center through 4G network, and the data center will detect it. As the object detection method for key components of transmission lines needs to be integrated into outdoor surveillance equipment, this method should be faster in low hardware configuration. YOLO5 is a rapid object detection model that can be applied to the actual working environment. In our application, we chose YOLOv5 as the basic model. According to our application characteristics, we improved the distance measurement in the K-means clustering, added an attention mechanism, and upgraded the loss function.

### *3.1. YOLOv5s model*

YOLO families are regression-based algorithms for object detection [22]. YOLOv5 is the fifth version of the YOLO series. YOLOv5 series contains four object detection versions: YOLOv5s, YOLOv5m (You Only Look Once Version 5 Middle), YOLOv5l (You Only Look Once Version 5

Large), and YOLOv5x (You Only Look Once Version 5 Extra Large). They have different network depths, network volumes, parameter quantities, and feature map widths [23–25]. Among them, YOLOv5s has the smallest network width and depth, so it is the fastest, but less accurate. Its network structure is shown in Figure 3.

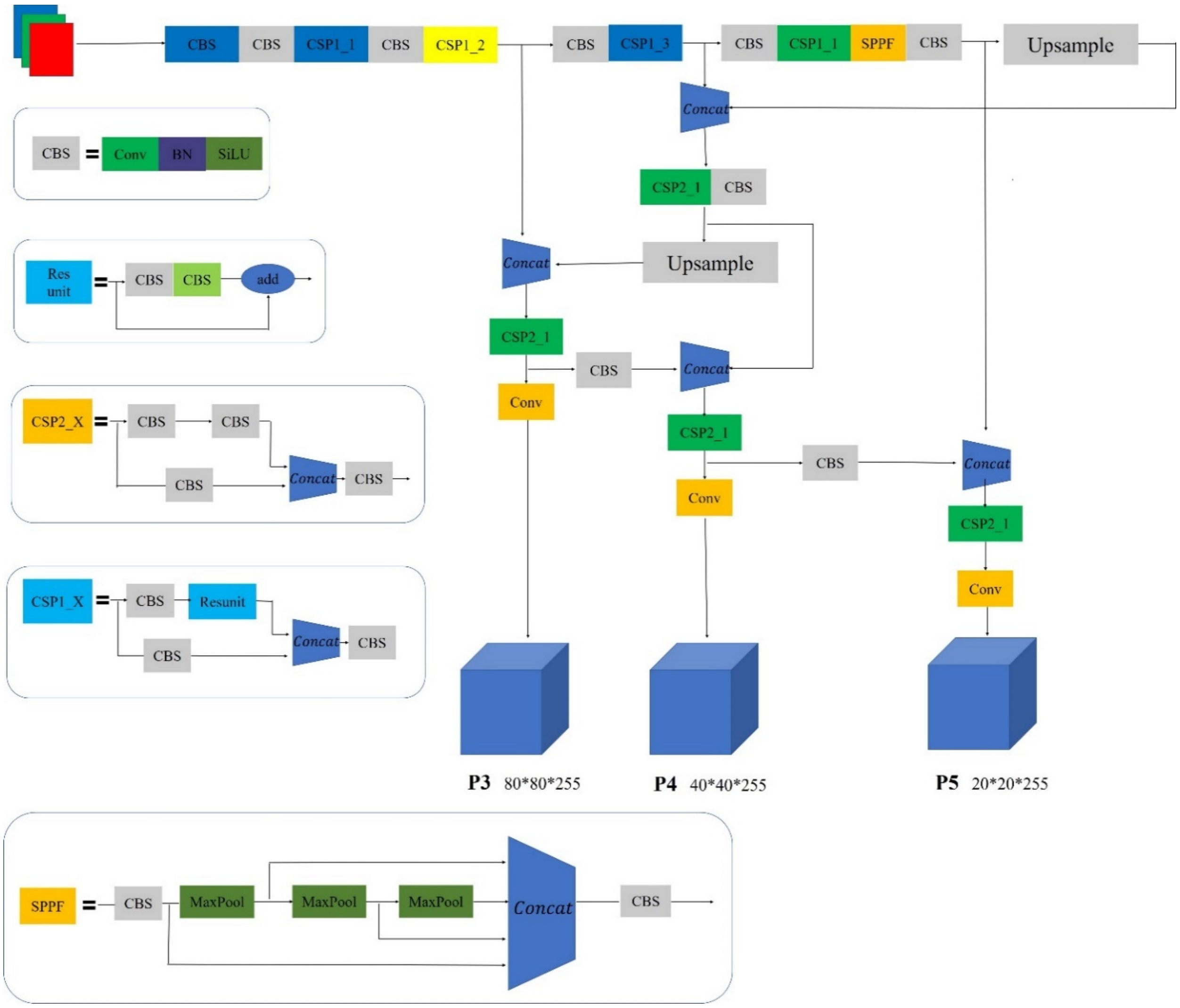

**Figure 3.** Yolov5s network structure.

### *3.2. Modify distance measurement in the K-means clustering*

In the YOLOv5s model, the K-means algorithm used Euclidean distance to measure sample distance. Since the poles and transmission towers are tall but narrow, the height-to-width ratio is large. Using the Euclidean distance to measure the distance between the predicted box and the real box of electric poles and transmission towers will lead to a significant error in the clustering results. To avoid this error, we considered replacing Euclidean distance with 1-IoU (Intersection over Union) distance. The 1-IoU distance can reduce this error caused by the large ratio of height to width of detected objects [26,27].

The 1-IoU distance $D_{IoU}$ based on IoU is calculated as follows:

$$D_{IoU} = 1 - IoU = 1 - \frac{|A \cap B|}{|A \cup B|} \tag{1}$$

where $A$ represents the real box and $B$ represents the prediction box.

Figure 4 shows the comparison of the Euclidean distance with the 1-IoU distance.

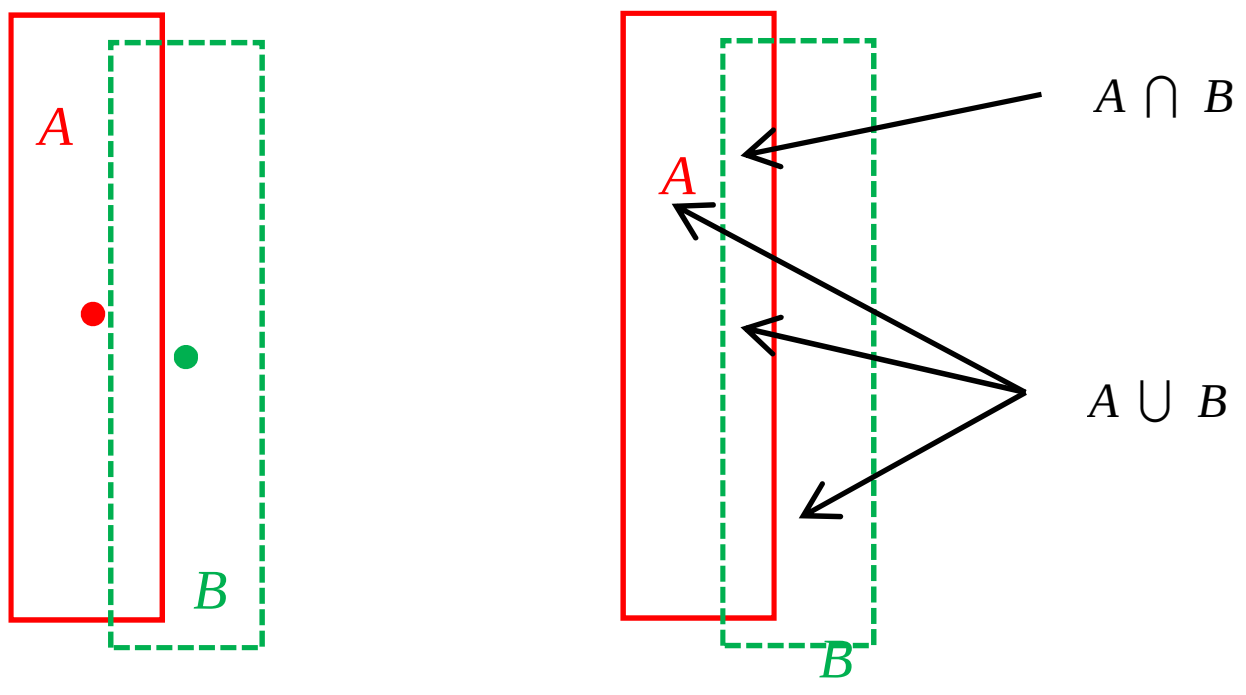

(a) Euclidean distance (b) 1-IoU distance

**Figure 4.** the comparison of the Euclidean distance with the 1-IoU distance.

When the height-to-width ratio is large, we compare the two distance metrics to measure the distance between a real box and a prediction box in Figure 4. The red rectangle A represents a real box, and the green rectangle B represents a prediction box.

In Figure 4(a), the Euclidean distance between the real box A and the prediction box B is calculated using the distance between the two boxes' centers. Although the positions of boxes A and B are very different, their centers are very near, so the Euclidean distance is minimal. In Figure 4(b), The 1-IoU distance between the real box A and the prediction box B is calculated using the intersection ratio between A and B. According to Eq (1), the overlapping area of A and B is small, so the IoU is small, and the 1-IoU distance is large. Therefore, if the detected targets significantly differ in width and height, the 1-IoU distance is better.

We use the 1-IoU distance in the K-means algorithm to reduce the error of anchor matching in the YOLOv5s model.

### *3.3. Adding attention mechanism*

Attention module has been proven to effectively enhance the representation ability of convolutional neural networks [28]. Many images have complex backgrounds and dense targets in the dataset of key components in the transmission line. It is necessary to improve its saliency to enhance the feature expression ability of the detected target in the complex background. Therefore, we introduce an attention mechanism to enhance features, mainly capturing the various iconic appearances of key components of transmission lines.

Attention mechanisms can be divided into two types: channel attention and spatial attention. Channel attention mainly explores the feature mapping relationship between different feature channels. Spatial attention uses multi-channel features in different spatial locations to build the relationship between two pairs, thus associating spatial context. The CBAM is an attention mechanism module that combines spatial and channel attention [35].

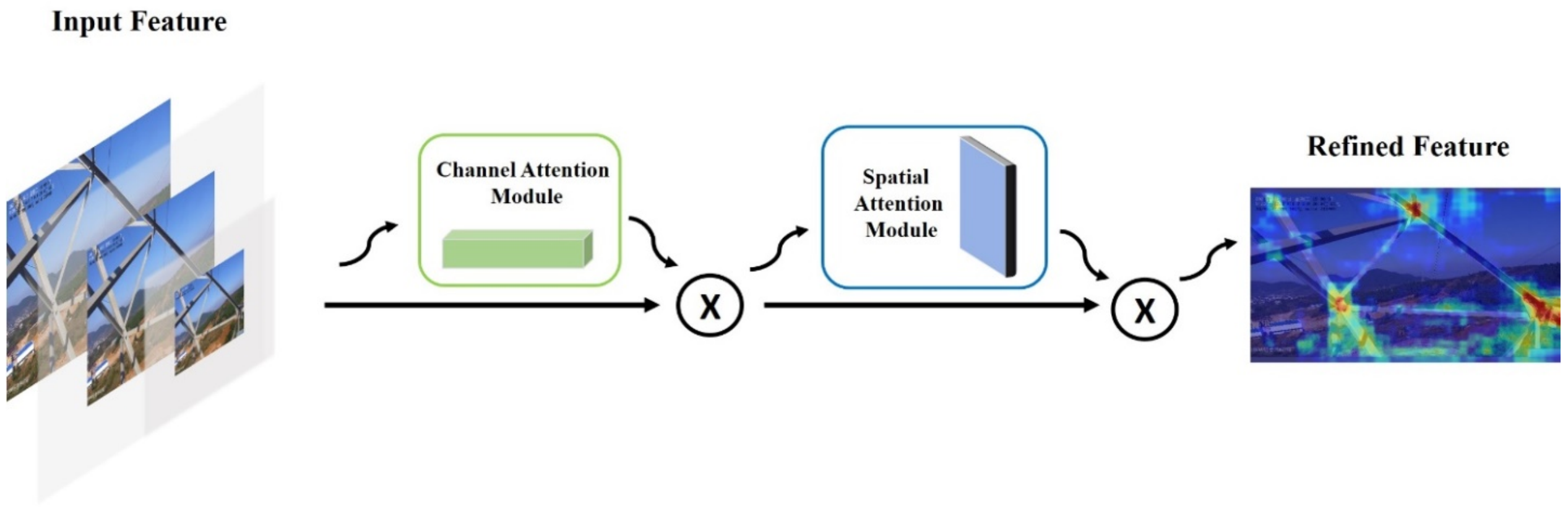

**Figure 5.** CBAM architecture for grid transmission line detection.

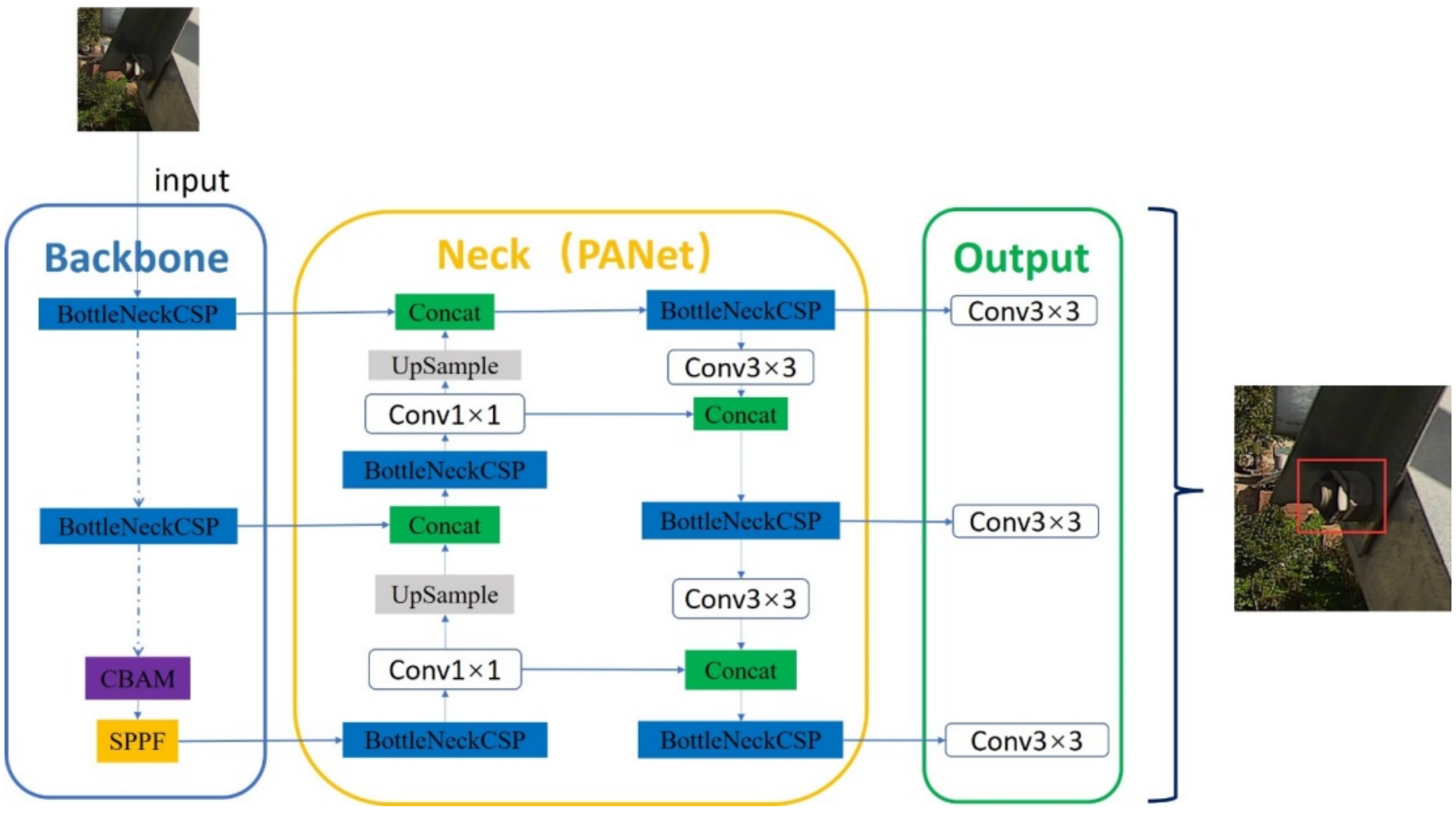

**Figure 6.** Our network structure after adding CBAM.

The CBAM automatically acquires the importance of each feature channel through learning. In addition, The CBAM automatically receives the significance of each feature space through similar learning methods and uses the obtained importance to enhance and suppress features that are not important to the current task. Our CBAM architecture for grid transmission line detection is shown in Figure 5.

Our network structure after adding the CBAM is shown in Figure 6. After adding the CBAM, our model can focus more on the detected target to reduce the classification error.

### *3.4. Focal loss*

The class imbalance problem typically occurs the instances of some classes are many more than others. The class imbalance usually affects the effectiveness of classification. In the dataset of transmission line key components, vehicles' samples are much smaller than other categories.

Focal loss is a loss function that can reduce the impact of class imbalance [29]. It was originally used to solve model performance caused by image class imbalance. It adds weights to positive and negative samples through a weight factor $a_t$, and adds weights to the corresponding loss of samples according to the difficulty of sample discrimination by adding a modulating factor $(1-p_t)^\gamma$. That is, add a smaller weight to the samples that are easy to discriminate, and add a larger weight to the samples that are difficult to discriminate.

The focal loss is calculated as follows:

$$FL(p_t) = -a_t(1-p_t)^\gamma \log(p_t) \tag{2}$$

where

$$p_t = \begin{cases} p, & y=1 \\ 1-p, & otherwise \end{cases} \tag{3}$$

where $p_t$ is the closeness to the ground truth (the class y). A large $p_t$ indicates that the closer to the class $y$, and a large $p_t$ means a high classification accurate. $\gamma$ is a controllable parameter, and $\gamma > 0$. $a_t$ is the shared weight of controlling the total loss of positive and negative samples.

## 4. Experiments

We gave the experimental results and analysis from several aspects of our model improvement, such as the distance measurement in K-means clustering, attention mechanism, and focal loss function. After that, we conducted ablation experiments and comparative experiments with other methods. Finally, we also performed detection experiments on another public insulator image dataset.

### *4.1. Data augmentation*

Due to the insufficient sample images, we used a data enhancement tool library (ImgAug) to augment our dataset. Data augmentation can expand training sets, improve the model's generalization ability, and effectively improve the robustness of the model.

The ImgAug library provides many image processing functions, which can efficiently realize the rotation, flipping, affine, brightness enhancement, contrast enhancement, color enhancement and other operations of the original image. The data augmentation effect of transmission line key components images using ImgAug is shown in Figure 7.

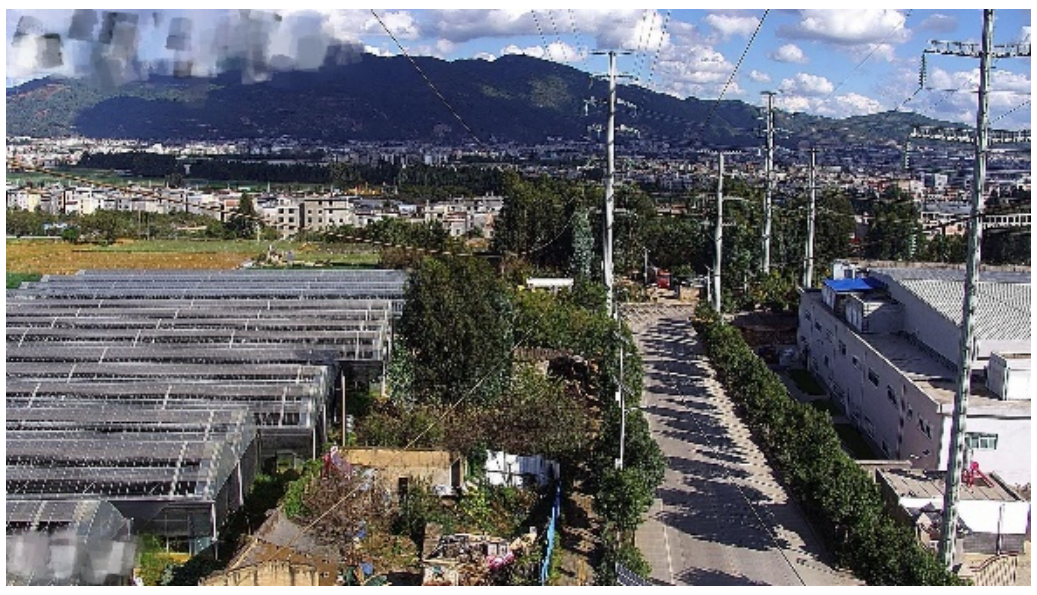

(a) Original image

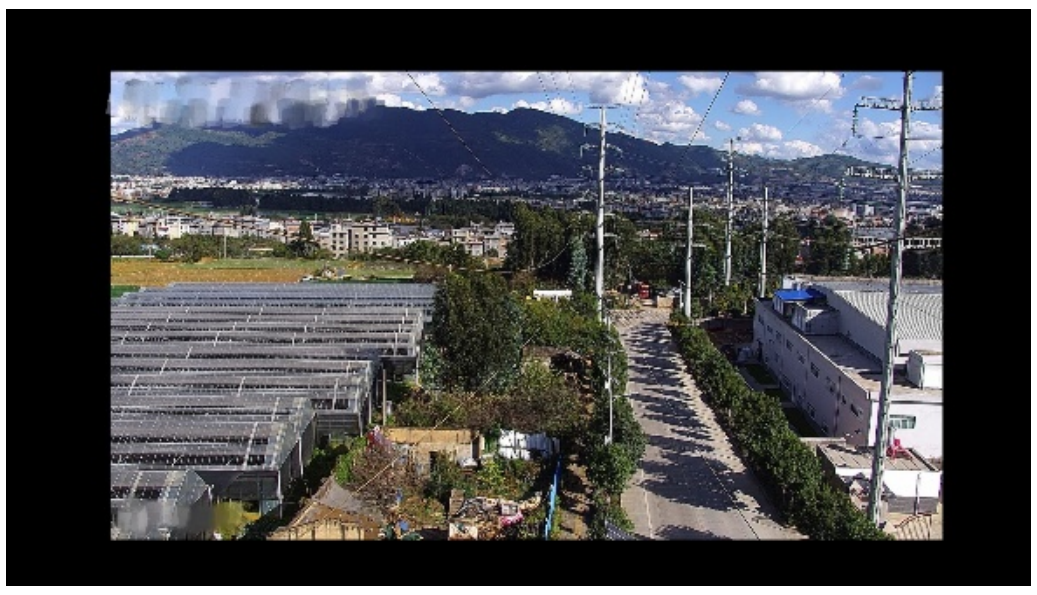

(b) Affine

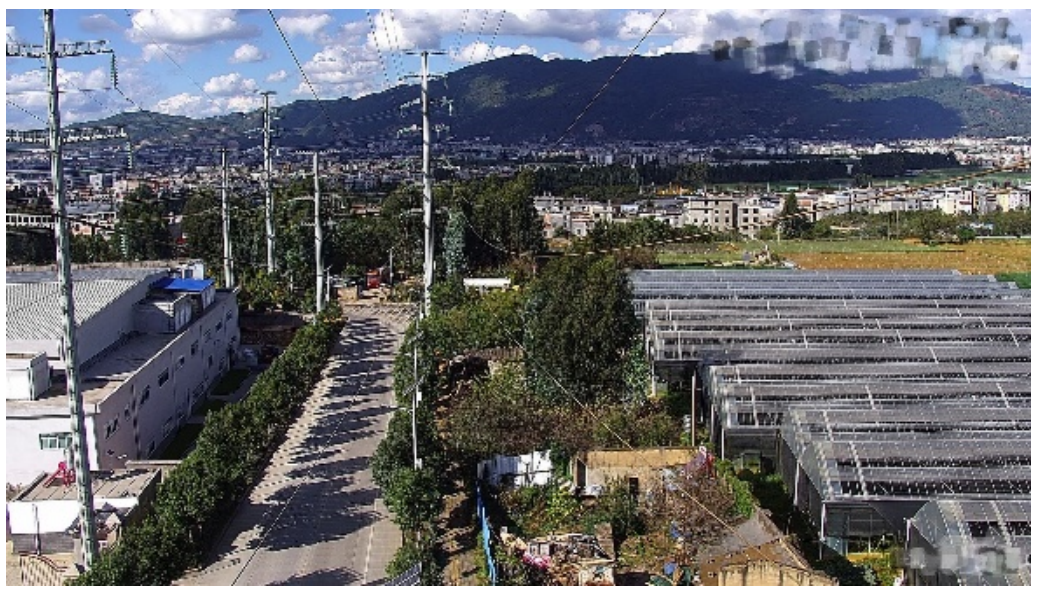

(c) Flip

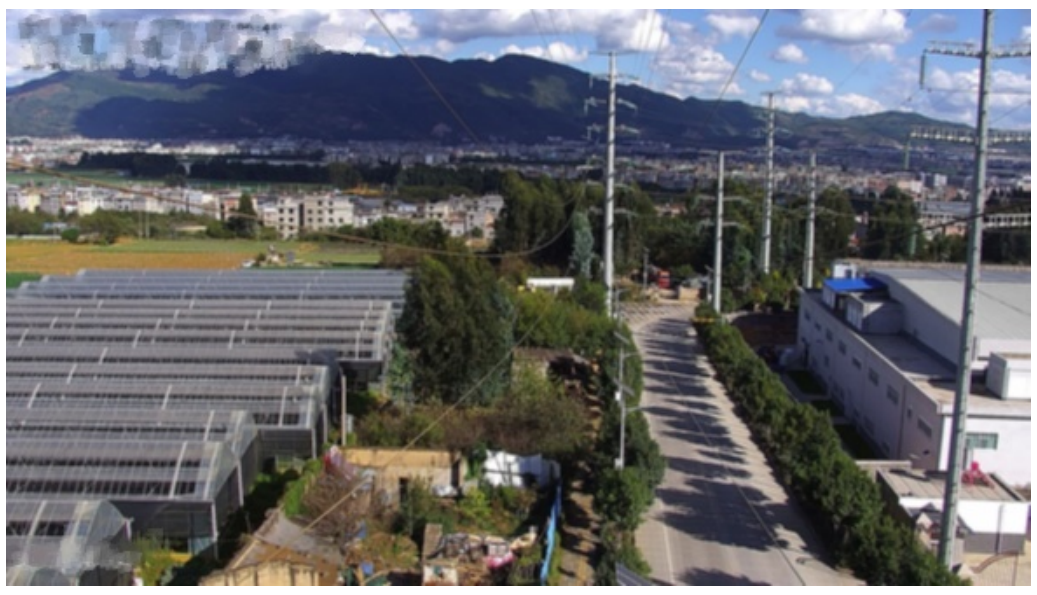

(d) GaussianBlur

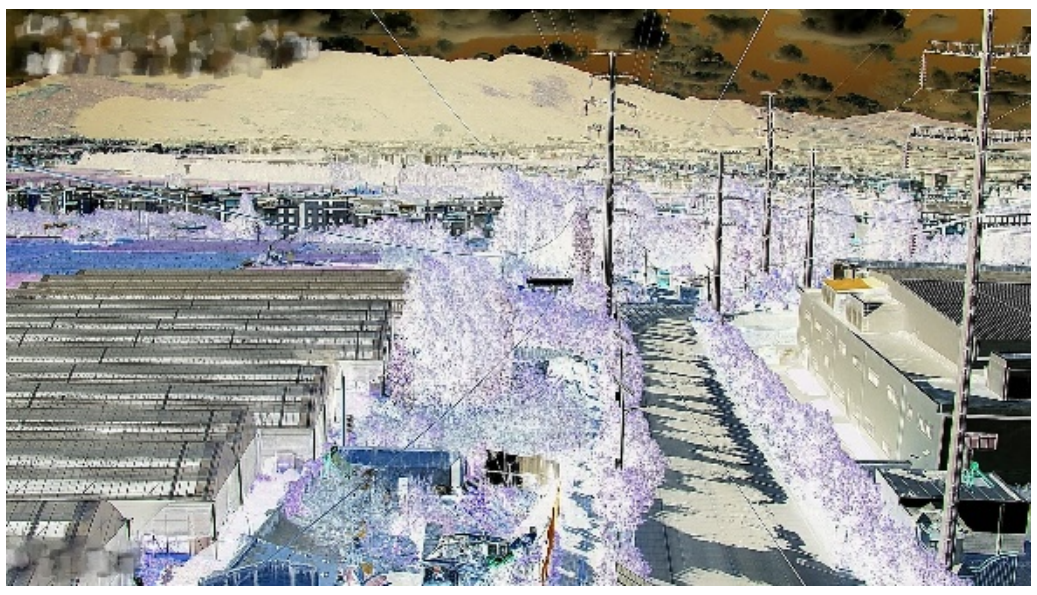

(e) Invert

**Figure 7.** Data augmentation effect using ImgAug.

There were 4290 images in our original dataset, and the number of images increased to 11,335 after data augmentation. We have labeled these detected targets. In many image samples, there may be multiple detected targets in an image sample. After counting the labeled targets, the number of five classes is shown in Figure 8.

We randomly divided the training set, validation set and test set in the ratio of 8:1:1, as shown in Table 1.

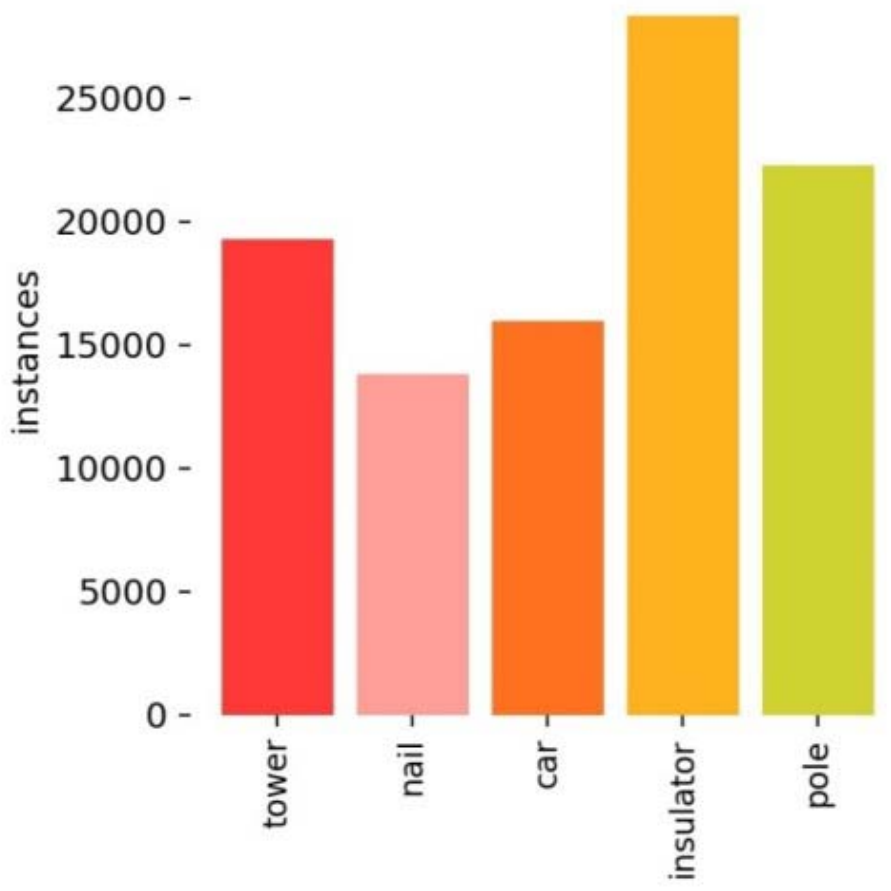

**Figure 8.** Number of samples in each category after data augmentation.

**Table 1.** Statistics of image dataset.

| Training set | Validation set | Test set | Total |
|---|---|---|---|
| 9067 | 1134 | 1134 | 11,335 |

*4.2. Experiment configuration and evaluation index*

We used Microsoft Windows 10 as the operating system, one NVIDIA GeForce GTX 2080Ti as the GPU and PyTorch 1.10.0 as the deep learning framework.

The learning rate momentum was set to 0.937, the batch size was 16, the initial learning rate was 0.01, the weight decay was 0.0005, and the training round was 300 to prevent overfitting.

The evaluation index used in the experiments were precision, recall and mean average precision (mAP). The precision ($P$) and recall ($R$) are as follows:

$$P = \frac{TP}{TP + FP} \quad (4)$$

$$R = \frac{TP}{TP + FN} \quad (5)$$

where, $TP$ is the number of samples that were positive and also correctly classified as positive; $FP$ is the number of samples that were negative but incorrectly classified as positive. $FN$ is the number of samples that were positive but classified as negative.

After obtaining the $P$ and $R$ of each category, a precision–recall (P-R) curve can be shown. $AP$ is represented by the area surrounded by the P-R curve and coordinates, and $mAP$ is the average of the $AP$ values of all categories. The $AP$ and $mAP$ are calculated as follows:

$$AP = \int_0^1 PR\mathrm{d}R \quad (6)$$

$$mAP = \frac{1}{N}\sum_{k=1}^{N} AP(k) \quad (7)$$

where $N$ represents the total number of categories, and $AP(k)$ represents the AP of the category $k$.

The speed index of the model is FPS (frames per second), and its reciprocal is the required time to process each image.

### *4.3. Experimental results and analysis*

#### 4.3.1. Experimental results of the distance measurement in K-means clustering

In Section 3.2, we modified the distance measurement in K-means clustering. We replaced the Euclidean distance with the 1-IoU distance. Our improvement can make the anchor boxes generated better adapted to the transmission line key components image datasets. There are three sets of preset anchors in YOLOv5s, each containing three anchors of different dimensions and shapes. The three sets of anchors are used to detect small objects in an 80 × 80 feature map, medium-sized objects in a 40 × 40 feature map, and large objects in a 20 × 20 feature map. Table 2 shows the anchor box sizes obtained by two distance measurement methods.

**Table 2.** Comparison of anchor box sizes in two distance measurement.

| Method | Small object | Medium object | Large object |
|---|---|---|---|
| Euclidean distance | [[6,6], [9,9], [7,17]] | [[13,28],[18,52], [32,53]] | [[45,62], [158,160], [182,199]] |
| 1-IoU distance | [[4,4], [7,7], [6,14]] | [[10,10],[15,8], [11,24]] | [[25,15], [17,48], [41,103]] |

As seen from Table 2, after using the 1-IoU distance in the K-means clustering, the height-width ratio of the anchor boxes for detected large objects becomes significantly larger. It can be more suitable for detecting transmission towers and poles in the dataset.

In the YOLOv5s model, the comparative experimental results of two distance measurement methods are shown in Table 3.

**Table 3.** Comparative experiments of two distance measurement.

| Method | mAP@0.5/% | Precision/% | Recall/% |
|---|---|---|---|
| Euclidean distance | 94.7 | 96.5 | **92.5** |
| 1-IoU distance | **95.3** | **97.1** | 91.5 |

As can be seen from Table 3, the mAP and accuracy have been improved after the 1-IoU distance is adopted. However, the recall decreases by 1%. To improve the model continuously, we added the CBAM attention mechanism module to upgrade the model's attention to the detected targets when the background is complex.

#### 4.3.2. Experimental results for adding the CBAM attention mechanism

To solve the low recall of the model, we added the attention mechanism module to the YOLOv5s model to improve the model's attention to essential features. We tried several standard attention modules: SENet (Squeeze-and-Excitation Networks) [32], ECA (Efficient Channel Attention) [33], CA (Coordinate Attention) [34] and CBAM [35].

Table 4 shows the comparison of each attention mechanism module.

**Table 4.** Comparison with the addition of each attention mechanism method.

| Method | mAP@0.5/% | Precision/% | Recall/% |
|---|---|---|---|
| YOLOv5s | 94.7 | 96.5 | 92.5 |
| YOLOv5s + 1-IoU | 95.3 | **97.1** | 91.5 |
| YOLOv5s + 1-IoU + SENet | 94.9 | 95.6 | 91.6 |
| YOLOv5s + 1-IoU + ECA | 94.9 | 95.1 | 91.6 |
| YOLOv5s + 1-IoU + CA | 95.3 | 93.7 | **93** |
| YOLOv5s + 1-IoU + CBAM | **95.5** | **97.1** | 91.8 |

Based on the three evaluation indexes in Table 4, the CBAM considers both accuracy and recall. Therefore, we chose to add the CBAM module to our improved model.

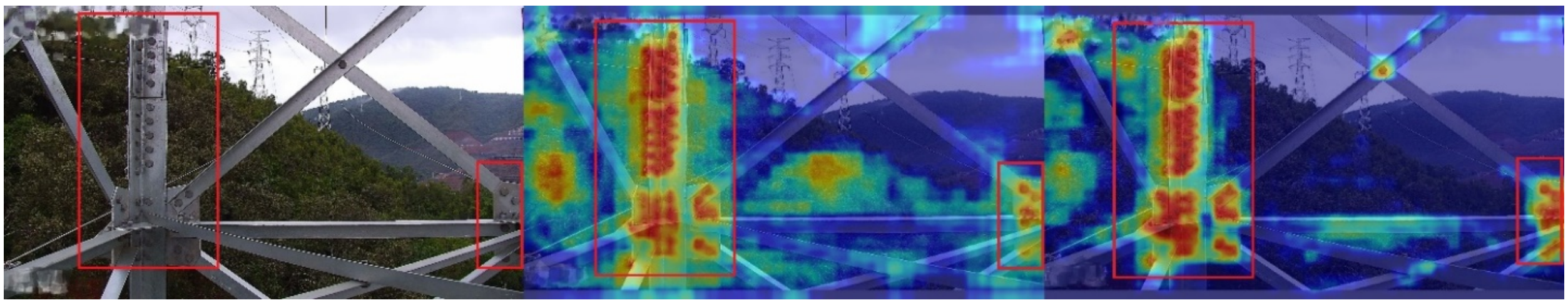

(a) Detection for dense small objects

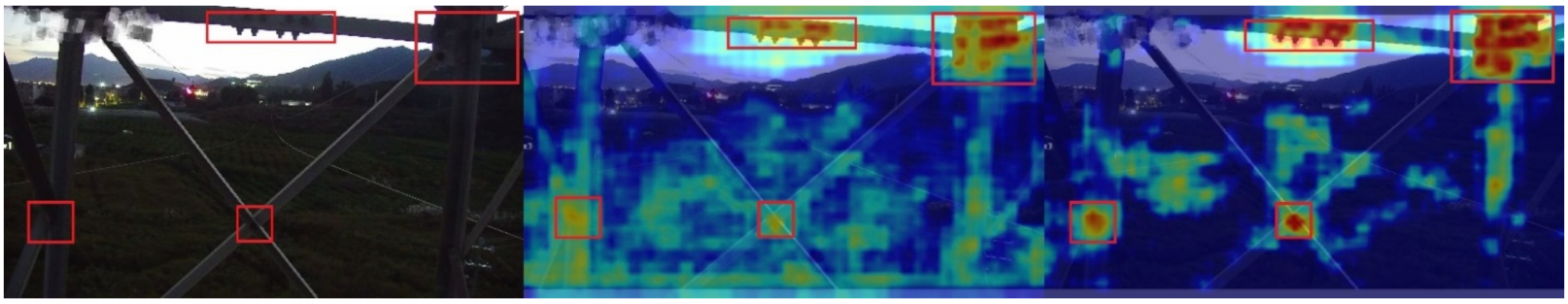

(b) Detection for blurred images

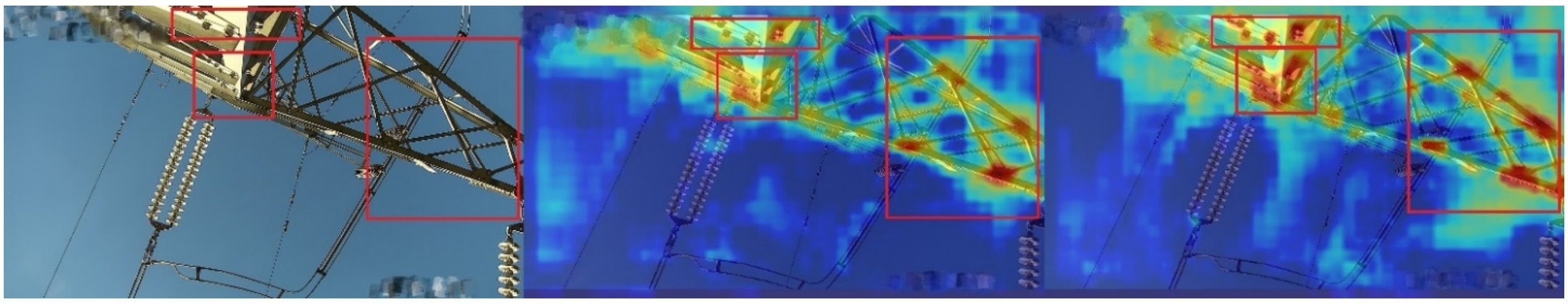

(c) Detection for the images shot with abnormal angles

*Continued on next page*

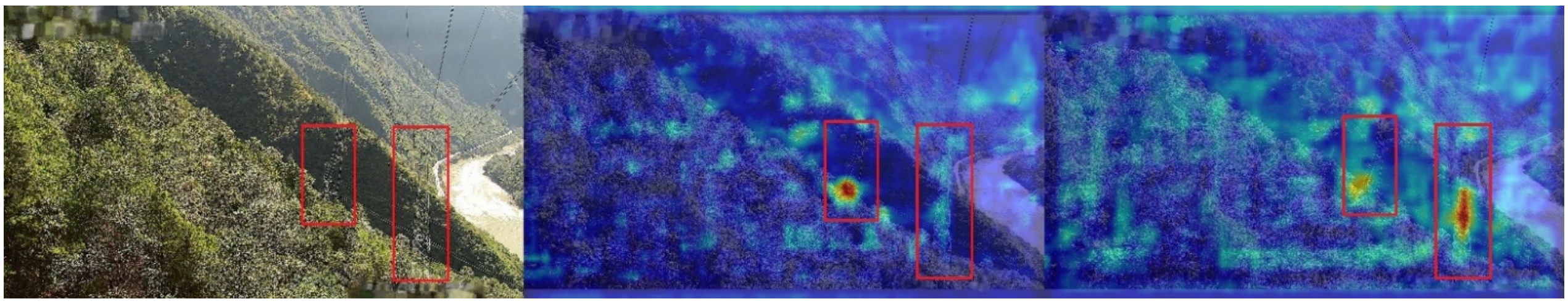

(d) Detection for complex background images

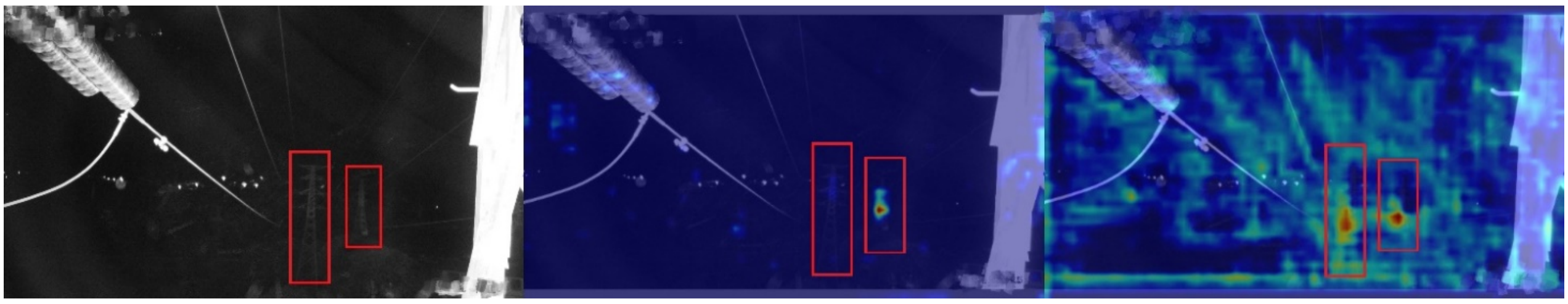

(e) Detection for the images shot in dark environments

**Figure 9.** Comparison results for adding the CBAM module.

Figure 9 shows the comparison results for adding the CBAM module into the backbone of the YOLOv5s model. The left images in Figure 9 are the original images, and the detected targets are in the red boxes. The center images show the detected target of the YOLOv5s model, and the right images show the detected targets after adding the CBAM module. In the middle and right images of Figure 9, the darker color represents the greater attention.

As seen in Figure 9, the CBAM module can enhance the saliency of the dense small targets and the targets in dark or complex backgrounds.

#### 4.3.3. Experimental results for the focal loss function

Although the class imbalance has been improved after our dataset augmentation, the number of screws and vehicles is still small in Section 4.1. In Table 4, the recall needs to be improved. After analyzing the experimental results, we found that the recall of screws and vehicles was too low. It is most likely due to the small sample images of these two categories.

Therefore, we used the focal loss function to reduce the impact of class imbalance. Table 5 shows the recall comparison using the focal loss function in each category.

**Table 5.** Recall comparison for using the focal loss function.

| Recall/% | Tower | Screws | Vehicle | Insulator | Pole | All |
|---|---|---|---|---|---|---|
| YOLOv5s | 93.6 | 94.8 | 78.9 | **100** | **95.2** | 92.5 |
| YOLOv5s + focal loss | **95.7** | **96.3** | 81.2 | 96.2 | 94.3 | 92.7 |
| YOLOv5s + Data Augmentation | 92.8 | 92.1 | 94.3 | 96.3 | 92.5 | 93.6 |
| Ours model + Data Augmentation | 92.9 | 94.7 | **96.3** | 96.5 | 91.5 | **94.4** |

As seen in Table 5, after our data augmentation and using the focal loss function, the recall has improved. In particular, the recall of vehicles has increased significantly.

Figure 10 shows the detected result comparison for vehicles.

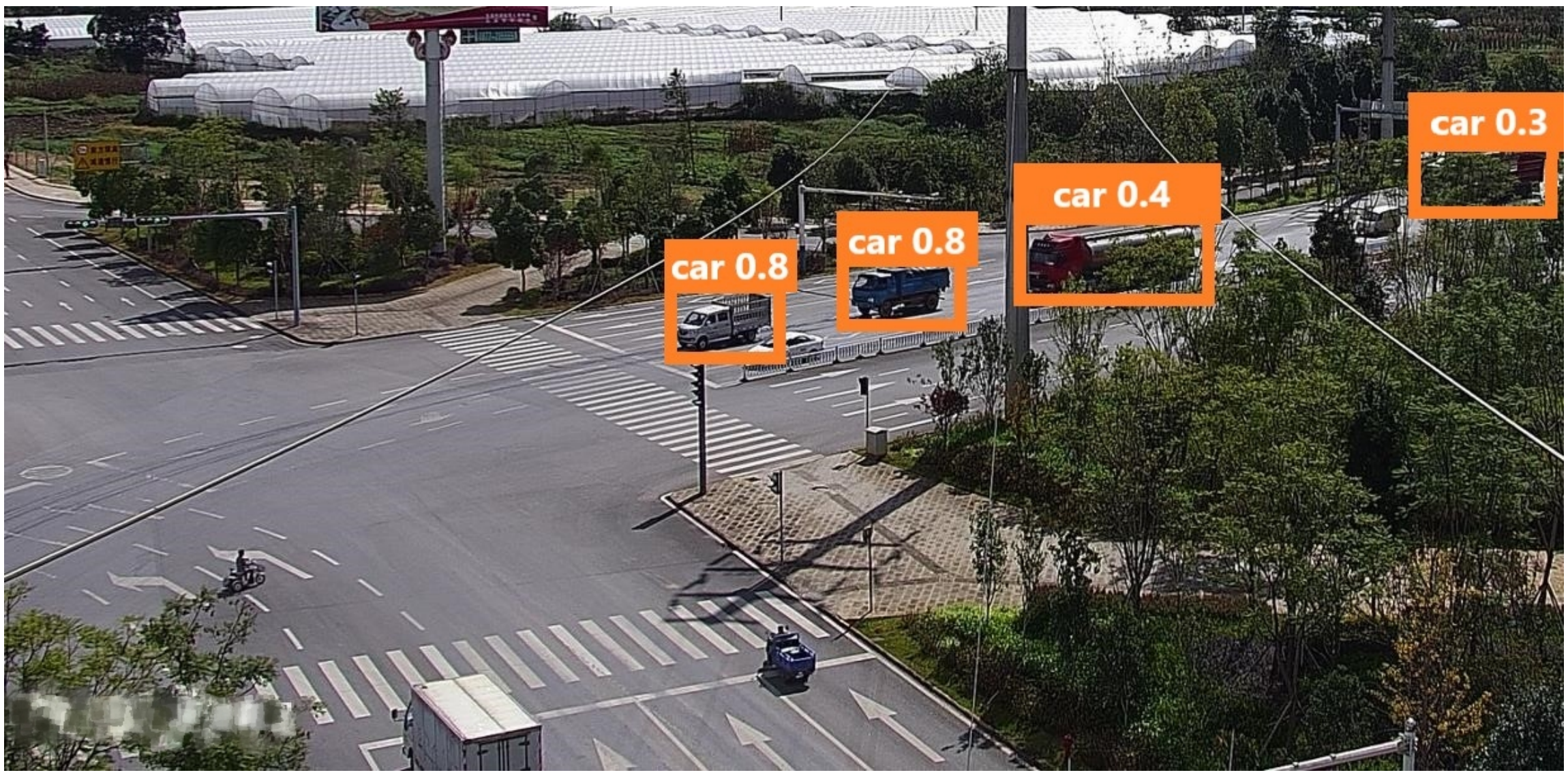

(a) Detection results of the YOLOv5s model

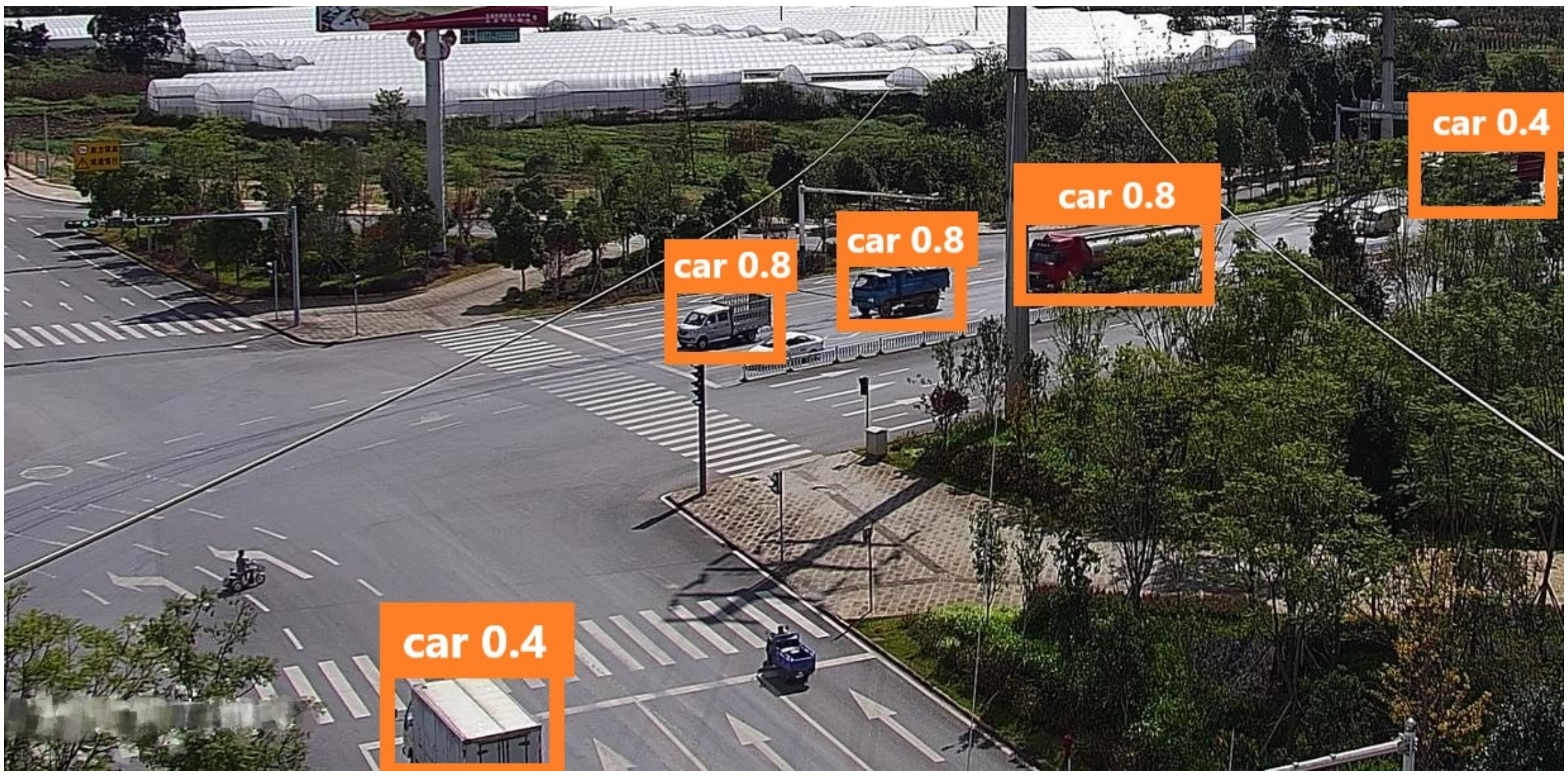

(b) Detection results of our improved model

**Figure 10.** Comparison for detecting vehicles.

In Figure 10(a), a car is missed, and the car is detected in Figure 10(b). In addition, the confidence of the detected cars is also higher in Figure 10(b). Therefore, our model reduces the missed detection of cars and improves the recall.

#### 4.3.4. Ablation experiment

We conducted more detailed ablation experiments to verify our improvements' effectiveness further. Table 6 shows the results of all ablation experiments.

**Table 6.** Ablation experiment results.

| CBAM | 1-IoU | Focal loss | Data augmentation | mAP@0.5/% | Precision/% | Recall/% |
|---|---|---|---|---|---|---|
| | | | | 94.7 | 96.5 | 92.5 |
| √ | | | | 95.6 | 96.7 | 92.6 |
| | √ | | | 95.3 | 97.1 | 91.5 |
| | | √ | | 94.7 | 96.2 | 92.7 |
| | | | √ | 96.1 | 97.0 | 93.6 |
| √ | √ | | | 95.5 | 97.1 | 91.8 |
| √ | | √ | | 94.7 | 94.1 | 92.5 |
| | √ | √ | | 95.1 | 96.4 | 92.6 |
| √ | √ | √ | | 95.9 | **97.9** | 92.9 |
| √ | √ | √ | √ | **98.1** | 97.5 | **94.4** |

As can be seen in Table 6, each of our improvements can improve the three evaluation indexes, especially the mAP. Finally, compared to the original YOLOv5s model, the mAP, precision, and recall enhanced by 3.4%, 1.0%, and 1.9%, respectively.

#### 4.3.5. Overall detection results

The comparison between our improved model and the original YOLOv5s model is shown in Figure 11.

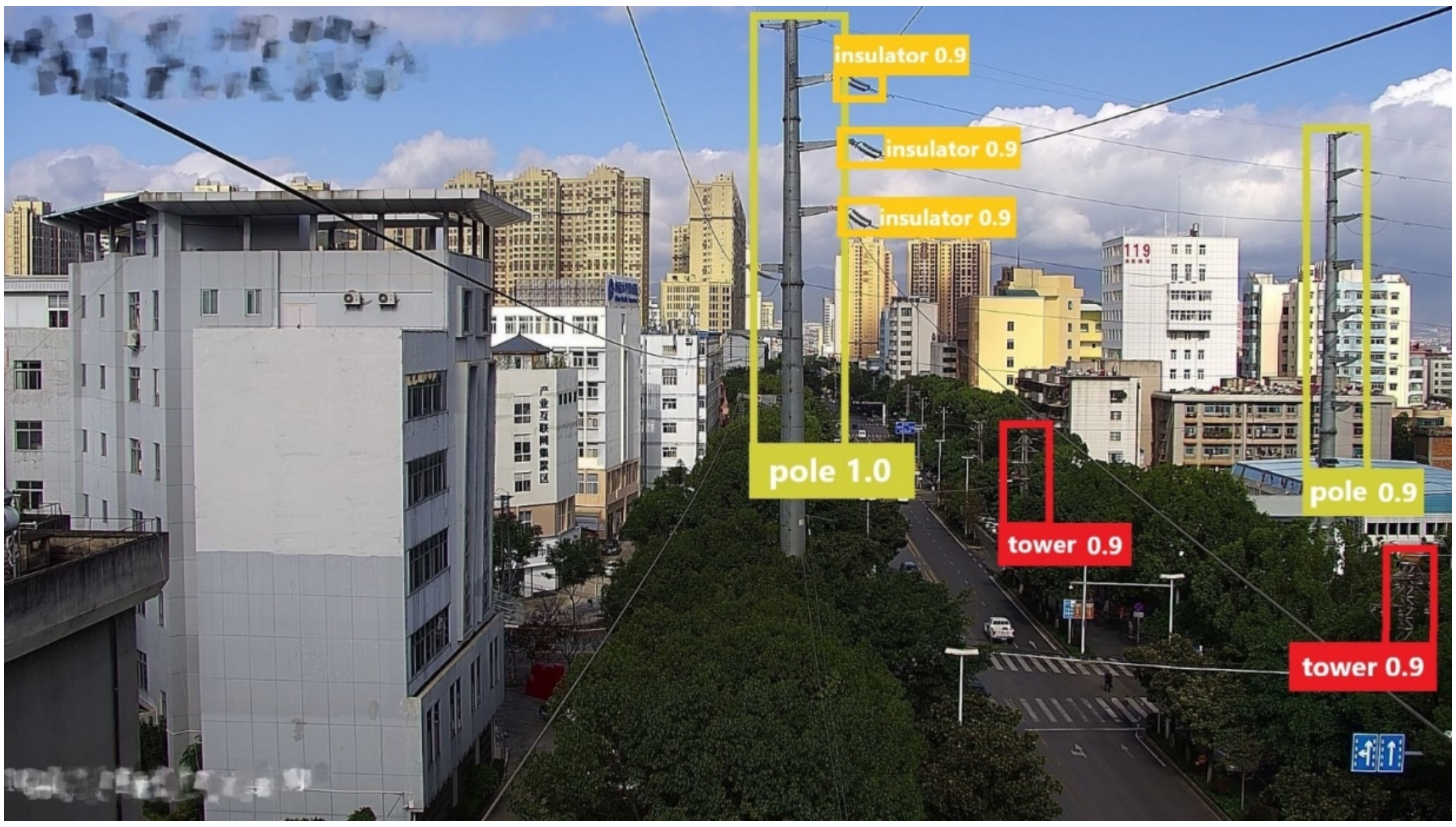

*Continued on next page*

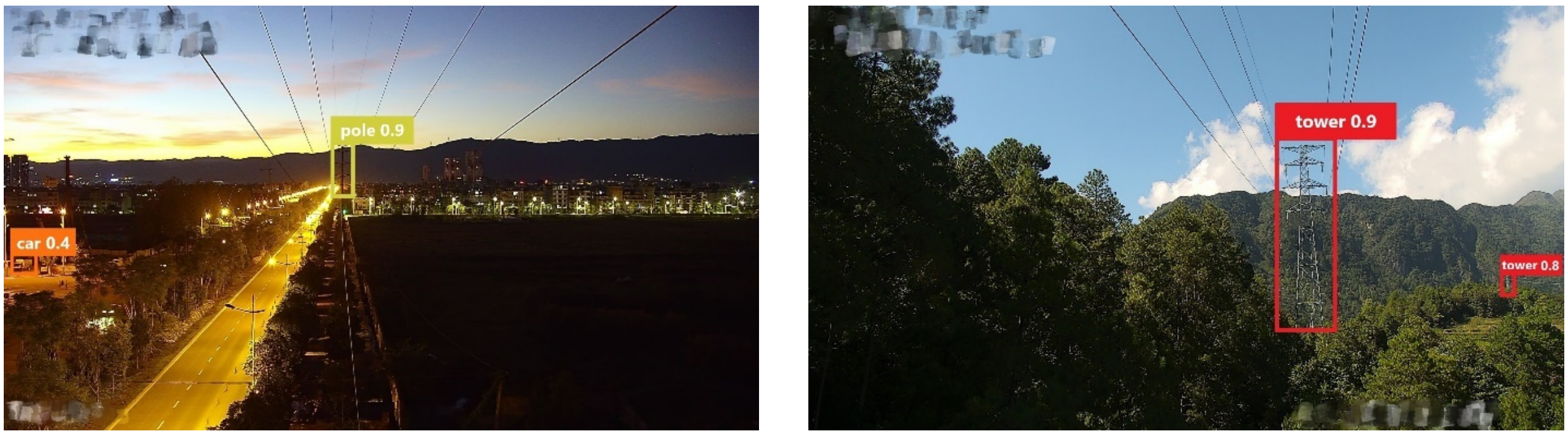

(a) Detection results of the original YOLOv5s model

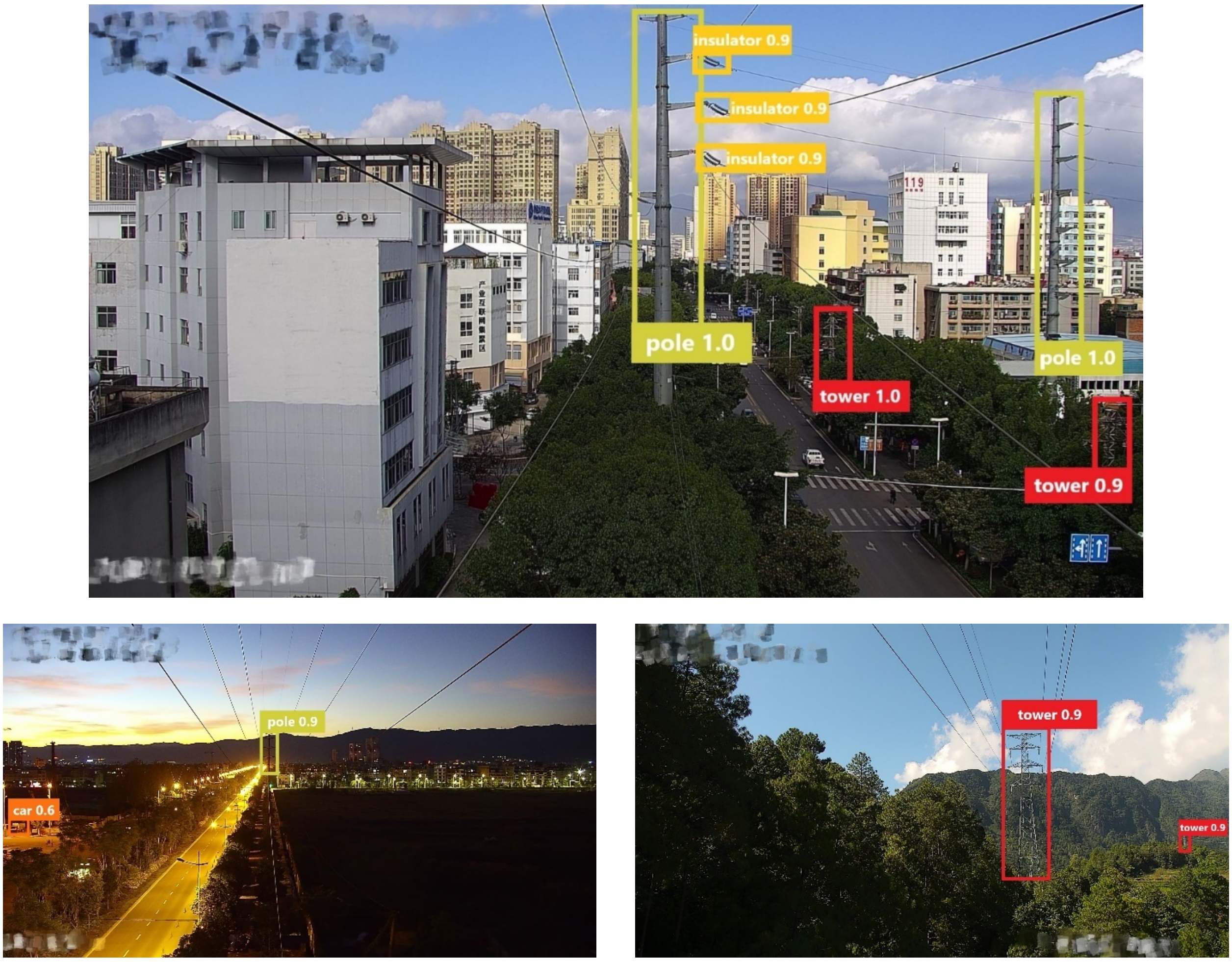

(b) Detection results of our improved model

**Figure 11.** Comparison between our improved model and the original YOLOv5s model.

It can be seen from Figure 11 that our improved model has higher confidence in detecting all categories, and there is no missing detection.

4.3.6. Comparison for model training

Figure 12 shows the loss change during the training for our improved model and the original

YOLOv5s model.

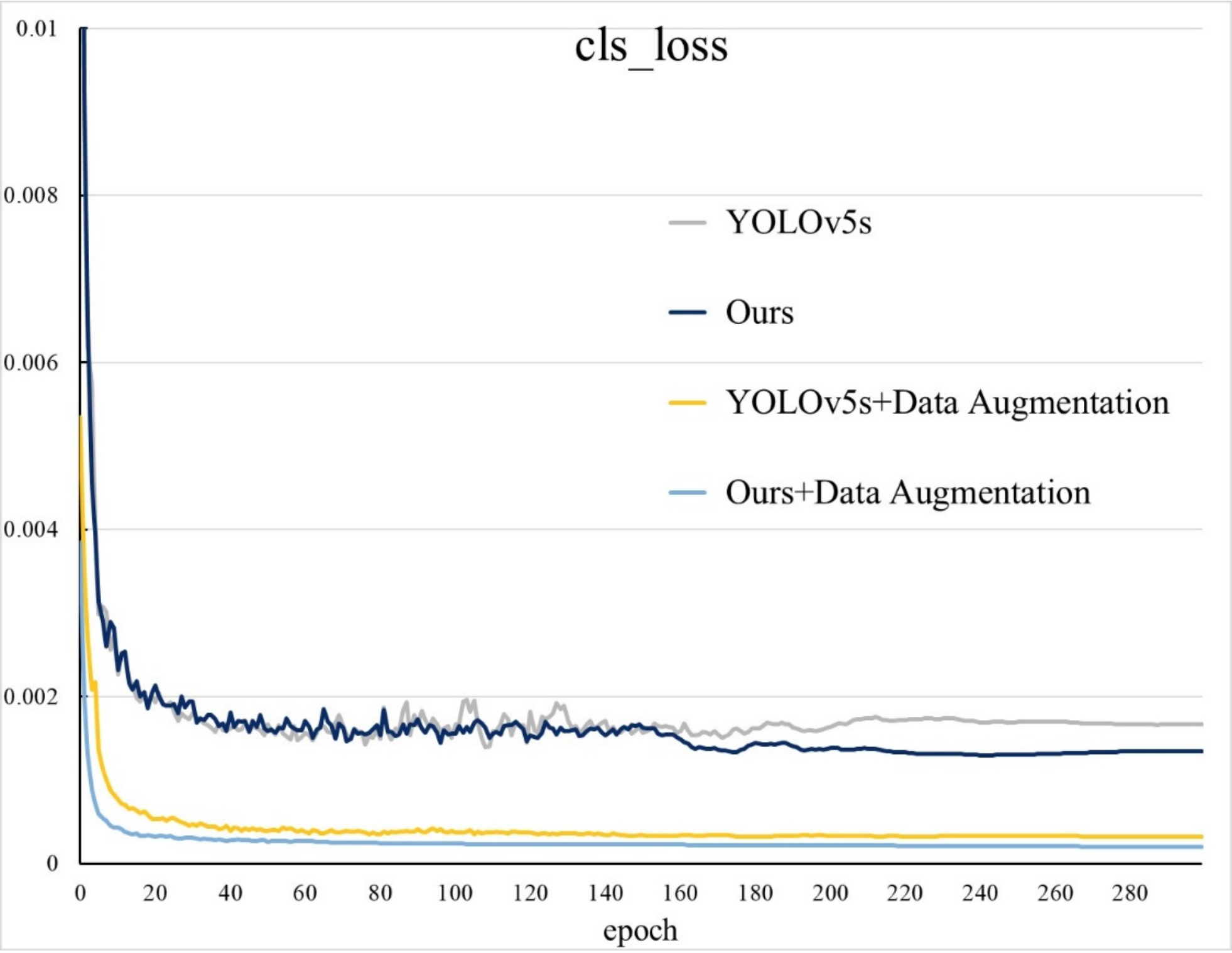

**Figure 12.** Comparison of loss curves.

In Figure 12, the gray curve is the loss curve of the original YOLOv5s model, and the dark blue curve is the loss curve of our improved model. The yellow curve is the loss curve of the original YOLOv5s model after data augmentation, and the light blue curve is the loss curve of our improved model after data augmentation. It can be seen that the loss values gradually stabilize as the training progresses. Regardless of the data augmentation, the loss value of our improved model is always smaller than that of the original YOLOv5s model. It indicates that our improved model has less loss and better convergence during training.

The mAP curves of our improved model and the original YOLOv5s model are compared in Figure 13. The mAP@0.5 indicates the average AP of each category when the IoU sets to 0.5. The mAP@0.5:0.95 represents the average mAP on the different IoU thresholds (from 0.5 to 0.95, in steps of 0.05).

As shown in Figure 13, the mAP@0.5 and mAP@0.5:0.95 of our improved model in this paper are higher than the original YOLOv5s model.

In Figure 13(a), the mAP@0.5 of our improved model eventually stabilizes at around 0.959 after iteration, while the mAP@0.5 of the original YOLOv5s model eventually stabilizes at around 0.947. Therefore, our improved model improves the mAP@0.5 by 1.2%. After our data augmentation, the mAP@0.5 of the original YOLOv5s model and our improved model rise to 96.1% and 98.1%, respectively, and it can be seen that the data augmentation effect is significant. After our data augmentation, the mAP@0.5 of our improved model is still higher than that of the original YOLOv5s model.

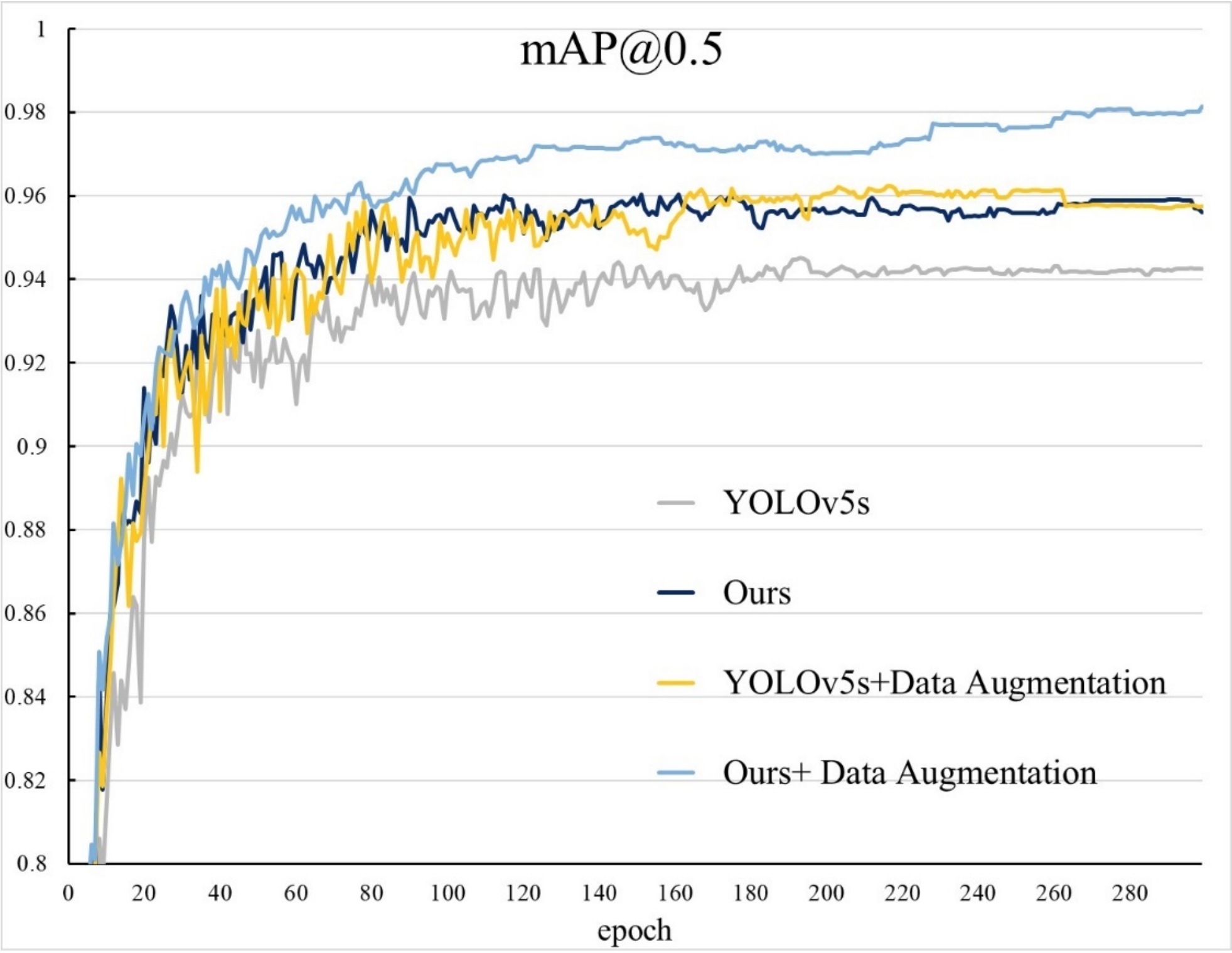

(a) Comparisons of mAP@0.5

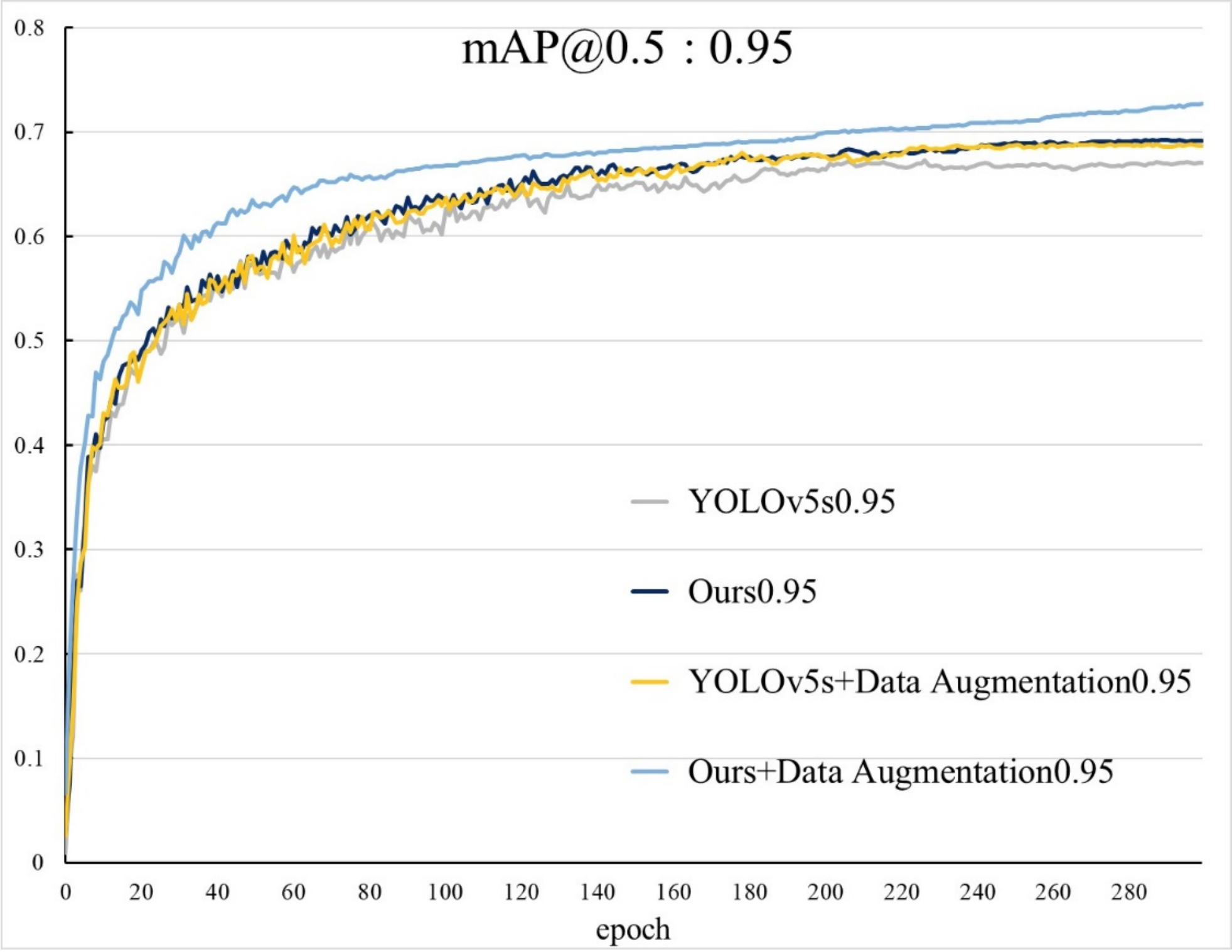

(b) Comparisons of mAP@0.5:0.95

**Figure 13.** Comparisons of mAP curve.

Similarly, in Figure 13(b), the four models' mAP@0.5:0.95 values also show that our improved model is adequate.

#### 4.3.7. Experimental comparisons with other models

We compared our improved model with the Faster R-CNN, SSD, CenterNet, YOLOv3, YOLOv4 and YOLOv5s models. We trained all object detection models using the same datasets with the same division and parameters, and the comparison experiment results are shown in Figure 14.

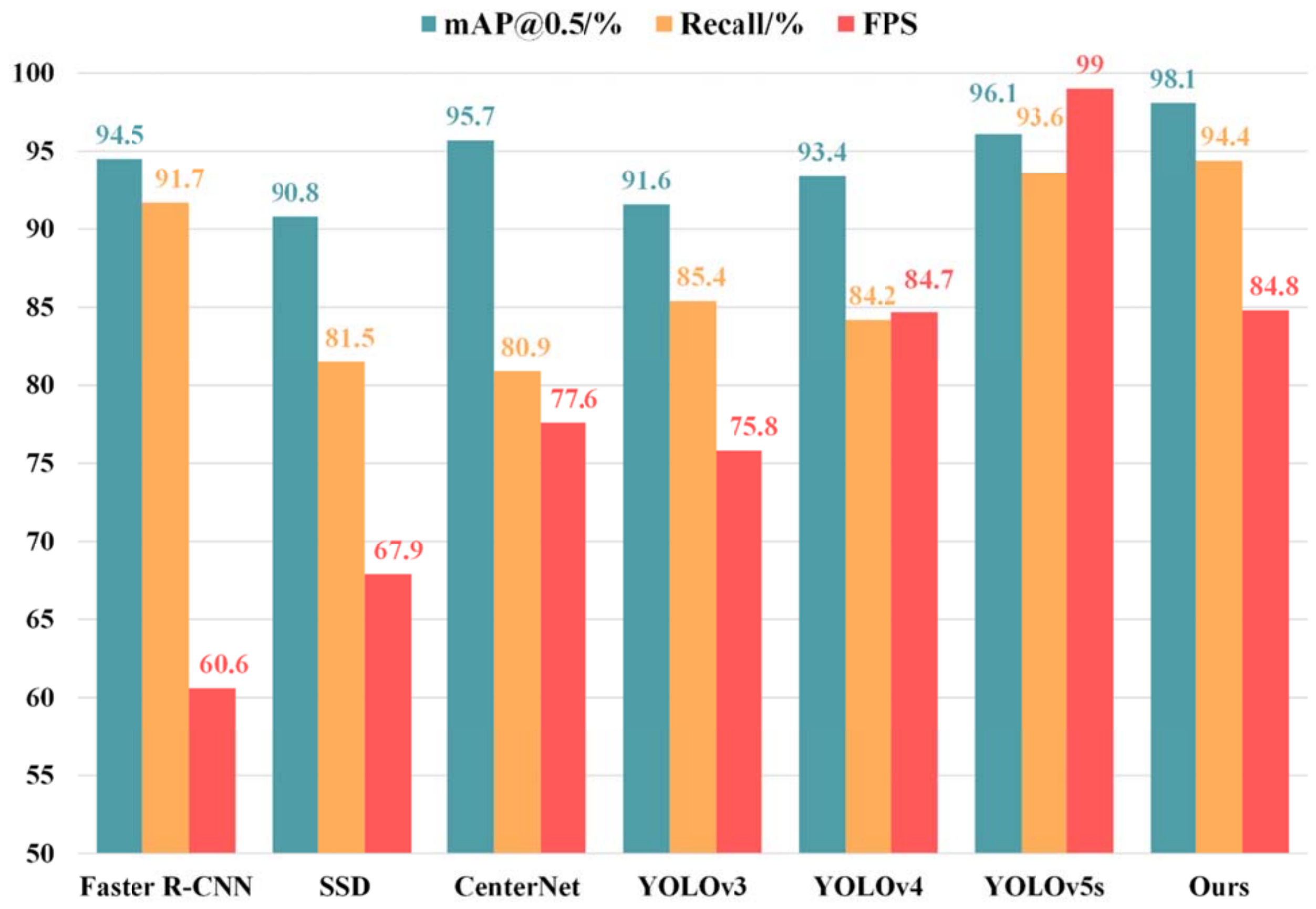

**Figure 14.** Experimental comparisons with other models.

Figure 14 shows that the mAP and recall of our improved model are higher than those of other models. Although our improved model's detection speed (FPS) is lower than the original YOLOv5s model, it is better than the other five models. Through comprehensive analysis, the experimental results prove that our improved model can balance the detection speed and accuracy.

#### 4.3.8. Experimental comparison on other public insulator image dataset

The reference [30] provided an available insulator image dataset and proposed a novel deep convolutional neural network (CNN) cascading architecture for localization and detecting defects in insulators. The dataset is 840 composite insulator aerial images collected by UAV, and each image has a resolution of 1152 × 864 pixels. The reference [31] also used this dataset.

In contrast experiments, we randomly divided the insulator dataset: 50% as the training set, 25% as the verification set, and the remaining 25% as the test set. We compared our improved model with the models in the references [30] and [31], and the results are shown in Table 7.

**Table 7.** Experimental comparisons in the remaining datasets.

| Method | mAP@0.5/% | Recall/% |
|---|---|---|
| CNN cascading architecture [30] | 91.0 | 96.0 |
| Attention mechanism + Fast RCNN [31] | 94.3 | 98.42 |
| Our model | **99.5** | **100** |

Table 7 shows that our improved model also performs better than the model of references [30] and [31] in the composite insulator dataset.

## 5. Conclusions

In this paper, we proposed an improved YOLOv5s model to meet the requirements of detecting key components of power transmission lines. Our model can automatically detect the key components of the transmission line, which is the preliminary work of the automatic transmission line inspection. The research work can reduce the workload and cost of transmission line inspection.

We modified the distance measurement in the K-means clustering, added the CBAM attention mechanism, and used the focal loss function. Our model improved the detection accuracy of key components of power transmission lines. The experimental results show that our improved model achieves 98.1% mAP@0.5, 97.5% precision, and 94.4% recall. However, the speed of our improved model is slightly slower than the YOLOv5s model.

Next, we will use fine-grained identification to detect the key components' defects and anomalies. We will refer to the references [36–38] for defect detection and abnormal detection of key components of power transmission lines to improve intelligent detection of power transmission lines.

## Conflict of interest

The authors declare there is no conflict of interest.